\documentclass[conference]{IEEEtran}
\usepackage{times}

\usepackage[numbers]{natbib}
\usepackage{multicol}
\usepackage[bookmarks=true]{hyperref}
\usepackage{amsmath}
\usepackage{amssymb}
\usepackage{graphicx} 
\usepackage{booktabs}
\usepackage{bbm}
\usepackage{multirow}
\usepackage{pifont}
\usepackage{amssymb}
\usepackage{overpic}
\usepackage{makecell}
\usepackage{tabularx}
\usepackage{ragged2e}
\newcolumntype{C}{>{\Centering\hspace{0pt}}X}
\usepackage[small]{caption}

\begin{document}

\title{Finding 3D Scene Analogies with Multimodal Foundation Models}

\author{Junho Kim\textsuperscript{1,2} and Young Min Kim\textsuperscript{1,2} \\
\textsuperscript{1}{\small Institute of New Media and Communications, Seoul National University} \\
\textsuperscript{2}{\small Dept. of Electrical and Computer Engineering, Seoul National University}
}

\maketitle

\begin{abstract}
Connecting current observations with prior experiences helps robots adapt and plan in new, unseen 3D environments.
Recently, 3D scene analogies have been proposed to connect two 3D scenes, which are smooth maps that align scene regions with common spatial relationships.
These maps enable detailed transfer of trajectories or waypoints, potentially supporting demonstration transfer for imitation learning or task plan transfer across scenes.
However, existing methods for the task require additional training and fixed object vocabularies.
In this work, we propose to use multimodal foundation models for finding 3D scene analogies in a zero-shot, open-vocabulary setting.
Central to our approach is a hybrid neural representation of scenes that consists of a sparse graph based on vision-language model features and a feature field derived from 3D shape foundation models.
3D scene analogies are then found in a coarse-to-fine manner, by first aligning the graph and refining the correspondence with feature fields.
Our method can establish accurate correspondences between complex scenes, and we showcase applications in trajectory and waypoint transfer.

\end{abstract}

\vspace{-0.5em}
\section{Introduction}
While no environment is completely identical to another, common patterns often exist in the spatial organization and layout of scene entities such as objects or floor corners.
Identifying such patterns, namely \textit{scene contexts}, and associating them with prior experiences is crucial for robots to plan and act in various unseen environments.
Recently, the 3D scene analogy task~\cite{3d_scene_analogies} has been proposed, where the goal is to find a smooth mapping in 3D space that links regions between two scenes sharing similar spatial context.
To illustrate, the 3D scene analogies shown in Figure~\ref{fig:qualitative} map points near the chair-table or sofa-cabinet group in one scene to the corresponding region in another.
Once found, these mappings can transfer motion trajectories or waypoints, which would be useful for planning or augmenting data for imitation learning~\cite{mimicgen}.

Existing approaches for 3D scene analogy estimation either extract and align neural descriptor fields or perform scene graph matching~\cite{3d_scene_analogies,sgaligner}.
For the former, one first trains a feedforward neural field that extracts descriptors for densely sampled query points within each scene, and finds a mapping that best aligns the descriptor field values.
While performant, this approach requires training descriptor fields specific to each target domain (e.g., indoor rooms), and generalization is not guaranteed outside the training data distribution.
The latter approach builds a scene graph with each object as a node and performs graph matching followed by ICP-based rigid alignment to find a dense mapping.
Here the graph matching process requires semantic labels to be known for accurate object association, which limits the applicability for open-vocabulary scenarios.

In this paper, we propose to use multimodal foundation models for finding 3D scene analogies.
By exploiting foundation models trained on large amounts of multimodal data~\cite{partfield,clip}, our method does not require additional training and can handle open-vocabulary setups.
Our method exploits a hybrid neural representation of sparse object-centric graphs and dense 3D fields for efficient scene analogy estimation in a coarse-to-fine manner.
Specifically, we build a graph storing each object as nodes along with their vision-language foundation model (CLIP~\cite{clip}) features and apply graph matching~\cite{rrvm_graph_match} between scenes to obtain a coarse object-level association.
The coarse associations are then refined to a smooth map by holistically aligning 3D shape foundation model (PartField~\cite{partfield}) features, which reduces dependence on individual features for robust scene analogy estimation amidst texture and shape variations.

Our approach based on multimodal foundation models can effectively find scene analogies in complex indoor scenes and is amenable to various downstream applications.
Quantitatively, our method outperforms existing approaches in mapping accuracy by accurately estimating scene analogies for complex indoor scenes.
Further, the dense maps found by our method can be used for transferring motion trajectories or waypoints, which indicates its potential for use in planning or data augmentation in imitation learning. 
We expect our work to serve as a practical pipeline based on foundation models for 3D scene analogy estimation.
\vspace{-0.5em}
\section{Method}
\begin{figure}[t]
  \centering
    \includegraphics[width=\linewidth]{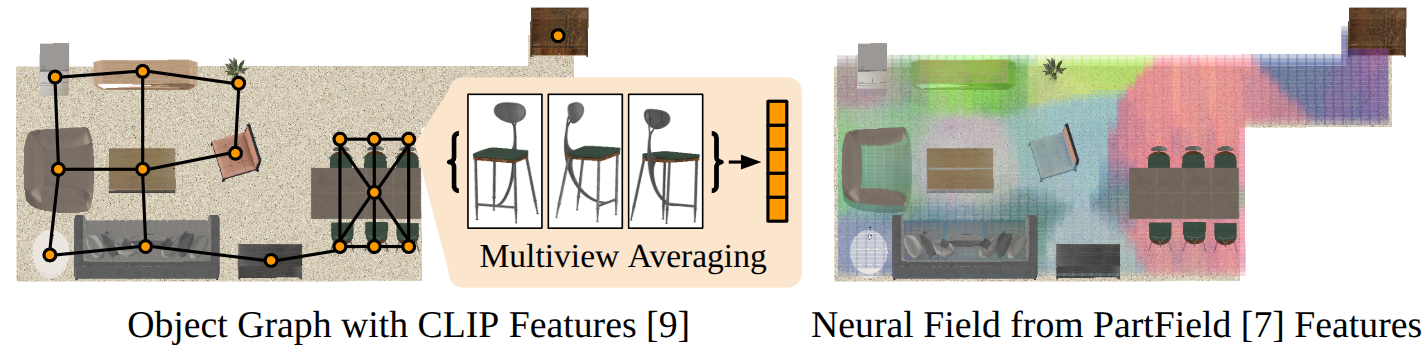}
    \vspace{-1.5em}
   \caption{Overview of our hybrid scene representation. Our method operates in a coarse-to-fine manner, by first obtaining instance-level associations from graph matching and refining the initial estimate with neural field alignment.}
   \label{fig:hybrid}
   \vspace{-2em}
\end{figure}
\subsection{Hybrid Scene Representation}
Given a pair of target and reference scenes $S_\text{tgt}$ and $S_\text{ref}$, our method finds a mapping $F(\cdot): S_\text{tgt} \rightarrow S_\text{ref}$ that transforms points in $S_\text{tgt}$ to corresponding points in $S_\text{ref}$ sharing similar scene contexts.
We express objects in both scenes as meshes, which can be obtained in practice from multi-view stereo~\cite{mvsnet} or dense visual SLAM~\cite{gsslam}.
Our method then builds a hybrid structure of sparse graphs and continuous feature fields.

\subsubsection{Graph Construction}
For each scene, we build a graph whose nodes contain object centroid coordinates and vision-language model (CLIP~\cite{clip}) features.
As shown in Figure~\ref{fig:hybrid}, the node features are extracted by rendering views around each object and averaging CLIP features extracted for each view.
Further, we connect edges for object centroid pairs whose distances are below a threshold (1.5m) and assign edge features as the average of adjacent nodes' CLIP features.

\subsubsection{Feature Field Extraction}
As shown in Figure~\ref{fig:hybrid}, we additionally construct a feature field $\Phi(\cdot): \mathbb{R}^3 \rightarrow \mathbb{R}^D$ from a 3D shape foundation model (PartField~\cite{partfield}).
The model takes as input a point cloud and outputs a feature vector for each point that captures the local geometry and part information.
We uniformly sample points from each object's mesh and extract PartField features.
The feature field for an arbitrary query point $\Phi(\mathbf{q})$ is then defined as the inverse distance-weighted interpolation~\cite{idw} of $k{=}100$ nearest PartField features.

\subsection{Coarse-to-Fine Scene Analogy Estimation}
Using the hybrid scene representation, our method finds scene analogies through a coarse-to-fine process.
First, our method applies graph matching to obtain coarse object-level associations.
Then, we cluster the object matches with DBSCAN~\cite{dbscan} and fit an affine map for each object cluster match.
Finally, for each object in the target scene $\mathcal{O}_i \subset S_\text{tgt}$ and its associated affine map $(\mathbf{A}_i, \mathbf{b}_i)$, we find optimal local displacements $\delta^*$ for each object point by minimizing the following cost function,
\begin{equation}
    C_\text{fine} = \sum_{\mathcal{O}_i \subset S_\text{tgt}}\sum_{\mathbf{p}_k \in \mathcal{O}_i} \|\Phi_\text{tgt}(\mathbf{p}_k) - \Phi_\text{ref}(\mathbf{A}_i\mathbf{p}_k + \mathbf{b}_i + \delta_k)\|_2,
\end{equation}
where the cost is defined for regularly sampled points from the object surface and $\Phi_\text{tgt}(\cdot), \Phi_\text{ref}(\cdot)$ are the feature fields for the target and reference scenes respectively.
The final mapping $F(\cdot)$ is found by fitting thin plate splines~\cite{tps} to all point-displacement pairs $\{(\mathbf{p}_k, \mathbf{A}_i\mathbf{p}_k + \mathbf{b}_i + \delta_k^*)\}_{i,k}$.

\section{Experiments}
\begin{figure}[t]
  \centering
    \includegraphics[width=\linewidth]{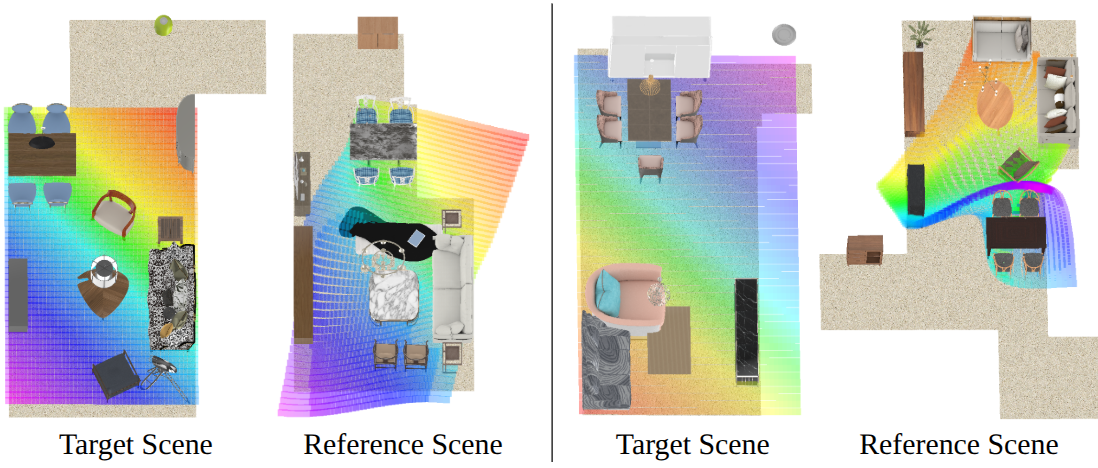}
    \vspace{-1.5em}
   \caption{Visualizations of estimated 3D scene analogies in 3D-FRONT [4]. We show mapping results for open-space points.}
   \vspace{-2em}
   \label{fig:qualitative}
\end{figure}
\subsection{Performance Analysis of 3D Scene Analogy Estimation}
We quantitatively evaluate our method using the 3D-FRONT~\cite{td_front} dataset, following Kim et al.~\cite{3d_scene_analogies}.
In Table~\ref{tab:quantitative}, we compare our method against neural contextual scene maps ~\cite{3d_scene_analogies} that align task-specific descriptor fields and scene graph matching~\cite{sgaligner} that perform object-level matching followed by ICP.
Our method outperforms the baselines on the Chamfer accuracy metric~\cite{3d_scene_analogies} which measures the percentage of target scene points whose distance to the nearest reference scene point is below a designated threshold.
By using multimodal foundation models, our method can perform accurate scene analogy estimation without additional task-specific training.
Figure~\ref{fig:qualitative} shows scene analogy visualizations, where our method reliably connects complex object layouts.

\begin{figure}[t]
  \centering
    \includegraphics[width=\linewidth]{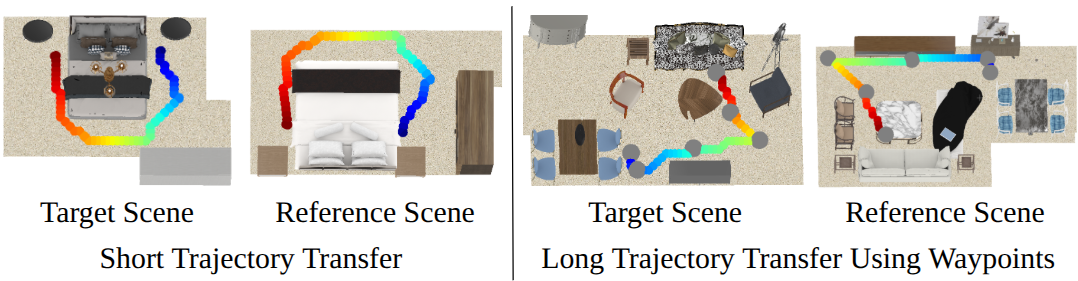}
    \vspace{-1.5em}
   \caption{Visualizations of trajectory transfer using 3D scene analogies.}
   \label{fig:trajectory_transfer}
   \vspace{-1em}
\end{figure}
\begin{table}[t]
    \centering
    \resizebox{0.7\linewidth}{!}{
    \begin{tabularx}{0.4\textwidth}{l|CCC}
    \toprule
    Metric & \multicolumn{3}{c}{Chamfer Acc.} \\ \midrule
    Threshold & 0.15 & 0.20 & 0.25 \\ \midrule
    Scene Graph Matching~\cite{sgaligner} & 0.13 & 0.35 & 0.44 \\
    Neural Contextual Scene Maps~\cite{3d_scene_analogies} & 0.55 & 0.69 & 0.72 \\
    Ours & \textbf{0.57} & \textbf{0.76} & \textbf{0.81} \\
    \bottomrule
    \end{tabularx}
    }
    \caption{Performance comparison with baselines.}
    \label{tab:quantitative}
    \vspace{-2em}
\end{table}
\subsection{Applications: Trajectory and Waypoint Transfer}
The smooth maps from our method can be applied to trajectory or waypoint transfer at scene-scale.
Such applications can aid in teleoperation~\cite{teleop}, planning~\cite{planning}, or data augmentation for imitation learning~\cite{mimicgen}.
To illustrate, suppose one has a long-horizon plan generated for a 3D scene using a simulator~\cite{isaac_lab,genesis_sim} and a task and motion planning (TAMP) algorithm~\cite{tamp}.
Using scene analogies, one can \emph{reuse} the plan found for one scene for solving a similar task in another, which will largely reduce the runtime and computational cost compared to planning from scratch.
To this end, one may first find a mapping from the original scene to the deployment scene, and transfer waypoints or continuous trajectories from the original plans to operate a robot in the deployment scene.

Figure~\ref{fig:trajectory_transfer} shows the trajectory and waypoint transfer results using our method.
Here we first find 3D scene analogies between the scene pairs and apply the estimated maps for transferring trajectories.
Our method can be applied flexibly depending on the length of the input trajectory.
First, given a short trajectory in the target scene, we estimate scene analogies and use the mapping to directly transfer each point in the trajectory to the reference scene.
For long trajectories, directly using the estimated maps may cause collisions.
Thus, we collect waypoints sampled from the trajectory and apply classical path planning~\cite{astar} on the mapped waypoints to generate the final long trajectory.
As shown in Figure~\ref{fig:trajectory_transfer}, our method can effectively transfer in both cases.
\section{Conclusion and Future Work}
In conclusion, we proposed a method for finding 3D scene analogies using multimodal foundation models.
We take a coarse-to-fine approach of first matching graphs defined over scene objects with vision-language model features, and refinement using 3D shape foundation models.
Initial results show that our method quantitatively outperforms existing methods and can transfer various trajectories and waypoints at scene scale.
Extending our work for efficient inference or handling dynamic objects is left as future work.

\IEEEpeerreviewmaketitle

\bibliographystyle{plainnat}
\bibliography{references}

\end{document}